\documentclass{article}

\usepackage{arxiv}
\usepackage{graphicx}
\usepackage{pythonhighlight}
\usepackage[utf8]{inputenc} 
\usepackage[T1]{fontenc}    
\usepackage{hyperref}       
\usepackage{url}            
\usepackage{booktabs}       
\usepackage{amsfonts}       
\usepackage{nicefrac}       
\usepackage{microtype}      
\usepackage{lipsum}
\usepackage{graphicx}
\graphicspath{ {./images/} }
\usepackage{float} 
\usepackage{subfigure} 

\title{ Gradient Policy on "CartPole" game and its' Expansibility to F1Tenth  Autonomous Vehicles}

\author{
  Mingwei Shi \\
  Second year  of Integrated Computer Science\\
  Trinity College Dublin\\
  College Green, Dublin 2\\
  \texttt{mshi@tcd.ie} \\
}

\begin{document}
\maketitle
\begin{abstract}

Policy gradient is an effective way to estimate continuous action on the environment. This paper, it about explaining the mathematical formula and code implementation. In the end, comparing between the rotation angle of the stick on CartPole , and the angle of the Autonomous vehicle when turning, and utilizing the $Bicycle $ $Model$, a simple Kinematics and dynamics model, are the purpose to discover the similarity between these two models, so as to facilitate the model transfer from CartPole to the F1tenth Autonomous vehicle\footnote{http://formulatrinity.com/the-car/}.

\end{abstract}


\section{Introduction}

Generally, when learners are now  studying the knowledge of the reinforcement learning algorithm at the beginning, the algorithm we first came up in learner's mind is the Q-learning algorithm, which is a classical reinforcement learning algorithm based on value iteration. In the state-to-action mapping process, an algorithm based on value iteration allows the system to explore in accordance with the policy guidelines, and update the state value at each step of the exploration.

Then, in value-based iteration, we have several problems that cannot prevent that. For example, when the value of each state is updated, it is necessary to estimate the probability of all actions. Unlike the discrete action of walking a maze, some cases such as robot control and automatic driving since the massive state information brought by continuous actions makes the calculation process almost impossible by tabular computation.

At this time, Policy Gradient, a reinforcement learning algorithm based on iteration policy, came into being. The policy gradient no longer calculates the reward, but directly calculates the probability of taking an action in a certain state, and directly selects the action through the probability.

Here is an example of David Silver in his deep reinforcement learning course $\footnote{https://www.davidsilver.uk/}$,
 To illustrate. In the maze shown below, there is now a robot to find gold coins. Among them, the reward for finding gold coins is +1, and the reward for encountering bombs is -1. In particular, the robot cannot distinguish between the two gray areas, that is, it cannot know whether it is in the gray area on the left or the gray area on the right.

\begin{figure}[H] 
\centering 
\includegraphics[width=0.7\textwidth]{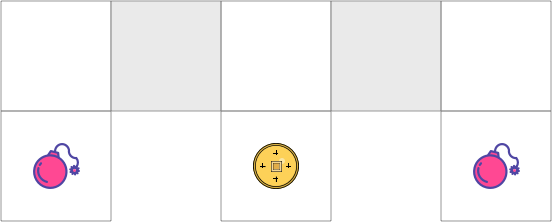} 
\label{Fig.main2} 
\end{figure}

If we adopt a value-based iterative method to learn, we will get a definite reward in a certain state. Therefore, the next action (left or right) of the gray (status) square is deterministic. That is, always left or right. This may lead to falling into the wrong cycle of the white grid on the left and the adjacent gray grid as showed in the figure below and failing to get the gold coins.

\begin{figure}[H] 
\centering 
\includegraphics[width=0.7\textwidth]{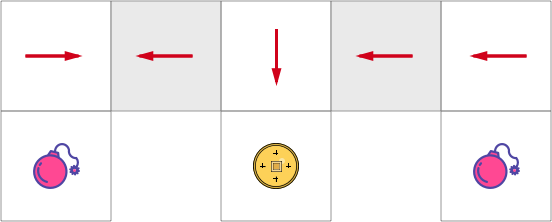} 
\label{Fig.main2} 
\end{figure}

When the strategy iteration method is used, the probability of the strategy output that the agent has learned, to move to the right and to the left is 0.5, so that it will not fall into the wrong cycle trap.

\begin{figure}[H] 
\centering 
\includegraphics[width=0.7\textwidth]{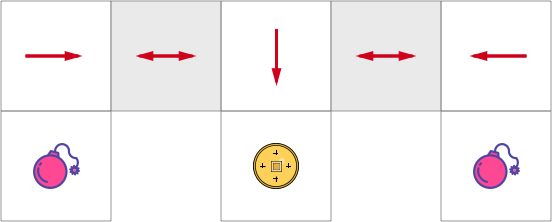} 
\label{Fig.main23} 
\end{figure}

This is the advantage of the strategy gradient method, which can handle continuous action scenarios where the value function cannot be applied. In addition, due to the probabilistic output, the problem of value function determinism is not applicable in certain scenarios.

\section{Mathematical Derivation of Policy Gradient Process}
\label{sec:headings}

In reinforcement learning based on value iteration, we complete learning by updating the value function:

$$   
Q_{\theta}(s, a)=f(\phi(s, a), \theta)
$$

Reinforcement learning based on strategy iteration directly completes the learning by optimizing the strategy function parameter $\theta$:
$$   
\pi_{\theta}(s, a)=g(\phi(s, a), \theta)
$$
The ultimate goal of learning based on strategy iteration is still the system to obtain the most rewards, in order to solve the optimal strategy function $\pi_\theta(s,a)$, then an objective function can also be used to measure the quality of the strategy. This is like the process of selecting the square loss function in supervised learning to measure the error between the true value and the predicted value, and then updating the parameters.

In general, according to different problem types, these three objective functions correspond to:
The first objective function $J_{1}(\theta)$ is suitable for reinforcement learning from a starting state each time

$$   
J_{1}(\theta)=V^{\pi_{\theta}}\left(s_{1}\right)
$$

Among them, $J_{1}(\theta)$ means that if the $Agent$ always starts to act from a certain state s1, or with a certain probability distribution from s1 to the end state to obtain cumulative rewards,
Also known as $Start Value$. And our goal is to maximize $J_{1}(\theta)$ .

Alternatively, you can use $Average Value$. For example, in a continuous environment where there is no starting state, taking into account the state distribution of the Agent at a certain moment, for each possible state calculation, it can continue to interact with the environment from that moment on. The rewards obtained are then summed according to the probability distribution of each state at that moment. For example, if you get 10 reward points in state 1, and 20 reward points in state 2, then the target reward value is 15 points.

The function $J_{a v V}(\theta)$ of $Average Value $ is as follow:

\begin{center}

$J_{a v V}(\theta)=\sum_{s} d^{\pi_{\theta}}(s) V^{\pi_{\theta}}(s)
$
\end{center}

In addition, you can also use another objective function: $Average reward per time-step$. In short, we can get the possibility that the $Agent$ is in all states within a certain time step, then calculate the rewards that can be obtained by taking all actions in each state, and finally sum all the rewards according to the probability distribution:

\begin{center}

$J_{a v R}(\theta)=\sum_{s} d^{\pi_{\theta}}(s) \sum_{a} \pi_{\theta}(s, a) R_{s, a}$
\end{center}

The $d^{\pi_{\theta}}(s)$ in the above two formulas is a normal distribution of the state of the Markov chain under the current strategy.

$J_{a v V}(\theta)$ doesn't want to get a cumulative reward result, but takes the average of the rewards available immediately and distributes it to the previous state. In fact, the above three objective functions achieve the same goal in different ways, all trying to get the value of the $Agent $at a certain moment.

With the value objective function, the next step is to maximize the objective function value through the optimization approaches, and get the corresponding parameter $\theta$ at the same time. Therefore, reinforcement learning based on strategy iteration actually returns to the optimization problem.

In the content of supervised learning, we use an algorithm called gradient descent to find the minimum value of the objective function. Nowadays, facing the problem of finding the maximum value of the objective function, it can also be done through the gradient, and the gradient ascent algorithm can be used here. Whether it is gradient descent or gradient ascent, it is actually a kind of thinking, but it faces the problem of minimum or maximum.

At this point, we let $J(\theta)$ be any type of strategy objective function, and the strategy gradient algorithm can make $J(\theta)$ rise to the local maximum along its gradient. At the same time, determine the parameter $theta $ corresponding to this local maximum value:

\begin{center}
    
 $\Delta \theta=\alpha \nabla_{\theta} J(\theta)$
\end{center}

\begin{center}
$\nabla_{\theta} J(\theta)=\left(\frac{\partial J(\theta)}{\partial \theta_{1}} \cdots \frac{\partial J(\theta)}{\partial \theta_{n}}\right)$
\end{center}

In the above formula, $\nabla_\theta $ is the policy gradient, and $\alpha $ corresponds to the learning rate.

For complex objective functions, the gradient is often inconvenient to calculate. Therefore, the gradient can be estimated by the finite difference method here. The principle of finite difference is roughly in the vicinity of the gradient, using Taylor expansion and retaining only the linear part, the formula is as follows:

\begin{center}
$\frac{\partial J(\theta)}{\partial \theta_{k}} \approx \frac{J\left(\theta+\epsilon u_{k}\right)-J(\theta)}{\epsilon}$
\end{center}

$\epsilon $ represents the step length, $u_ku $
  Is the unit vector. The use of the finite difference method does not require that the strategy function can be differentiated, and there is no need to calculate the gradient, which is very convenient. However, this method is a rough estimate since the drawback of this approach often exists noise, which is inefficient.

Since finite difference is not a better method, then we have to use the method of calculating gradient to update the strategy, which requires the strategy function to be differentiable. The concept of $Likelihood ratios $ is used here, that is, the gradient of the function at a certain variable $ \theta $is equal to the product of the function value and the gradient of the function's natural logarithmic function here:

\begin{center}

$\nabla_{\theta} \pi_{\theta}(s, a)=\pi_{\theta}(s, a) \frac{\nabla \pi_{\theta}(s, a)}{\pi_{\theta}(s, a)}=$
\end{center}

\begin{center}

$\pi_{\theta}(s, a) \nabla \log \left(\pi_{\theta}(s, a)\right)$
\end{center}

Among them, we also call $\nabla \ln \left(\pi_{\theta}(s, a)\right)$
It is the Score function. The score function function has a better property, that is, the logarithmic function can turn multiplication into addition to facilitate derivation.

For example, when using the well-known $ Softmax function $and linear function to construct the strategy function:

\begin{center}

$\pi_{\theta}(s, a)=\frac{e^{\phi(s, a)^{T} \theta}}{\sum_{b} e^{\phi(s, b)^{T} \theta}}$
\end{center}
At the same time,the $ Score function $ is:

\begin{center}
$
=\phi(s, a)^{T}-\mathbb{E}_{\pi_{\theta}}[\phi(s, \cdot)]
$
\end{center}

Or in a continuous space, assuming that the action $a$ conforms to the Gaussian distribution, the strategy uses a Gaussian distribution with a standard deviation of $1$. This strategy is called a Gaussian strategy. The scoring function of the Gaussian strategy is:

\paragraph{Policy gradient theorem.} 

We approach this task as a regression problem. For every item and shop pair, we need to predict its next month sales(a number).

Having said so much, in fact our goal is to solve $\nabla {\theta}J\left( \theta \right)$. Well, for any differentiable strategy. Then, for any differentiable strategy :$\pi _{\theta}\left( s,a \right)$
and the objective function of any strategy, and the objective function of any strategy
$J_1(\theta)$, $J{avV}( \theta)$, $J_{avR}(\theta)$, 
all have:
\begin{center}
$\nabla_{\theta} J(\theta)=\mathbb{E}{\pi \theta}\left[\nabla{\theta} \log \pi_{\theta}(s, a) Q^{\pi_{\theta}}(s, a)\right]$

\end{center}

The above formula is also called the strategy gradient theorem. The strategy gradient theorem reveals the consistency of different objective functions to solve the gradient. Knowing how to calculate the gradient, then the parameters can be solved by gradient ascent.
\paragraph{Monte Carlo Policy Gradient.}
In the Sales train dataset, it only provides the sale within one day, but we need to predict the sale of next month. So we sum the day's sale into month's sale group by item, shop, date(within a month).
In the Sales train dataset, it only contains two columns(item id and shop id). Because we need to provide the sales of next month, we add a date column for it, which stand for the date information of next month.

In order to solve the policy gradient optimization problem, we need to calculate $\nabla_{\theta} \log \pi_{\theta}(s, a)$ and $Q^{\pi_{\theta}}(s, a)$. According to the above content, we can find $\nabla_{\theta} \log \pi_{\theta}(s, a)$,
And how can the value of $Q^{\pi_{\theta}}(s, a)$ be obtained?

$Monte-Carlo Policy Gradient $ allows the system to generate a state sequence from the starting state to the ending  :$State -> Action ->Reward $ :

\begin{center}
$s_{1}, a_{1}, r_{2}, \cdots, s_{T-1}, a_{T-1}, r_{T} \sim \pi_{\theta}$
\end{center}

Next, the algorithm initializes the parameter $\theta$ randomly, and uses the value of $v_{t}$ from $t=1$ to $t=T-1$ as $Q^{\pi_{\theta}}\left(s_{t}, a_{t}\right)$  to solve the policy gradient optimization problem.

The pseudo code of the Monte Carlo policy gradient  is as follows:

\begin{figure}[H] 
\centering 
\includegraphics[width=0.7\textwidth]{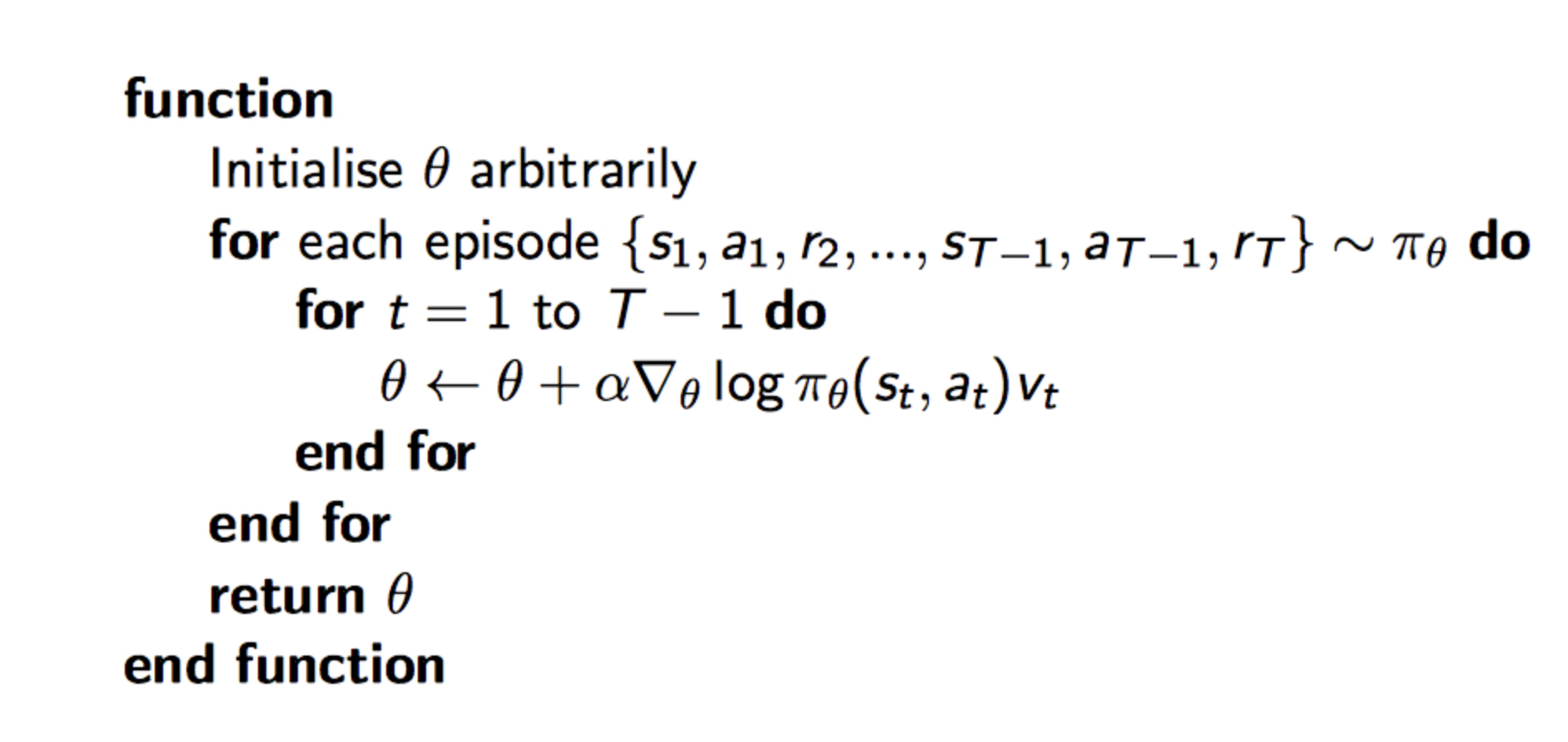} 
\label{Fig.main2} 
\end{figure}

\paragraph{Actor-Critic Policy Gradient}

In addition to the Monte Carlo strategy gradient algorithm, there is also a strategy gradient algorithm called $Actor-Critic$. A big flaw of the Monte Carlo strategy gradient algorithm is that the variance is relatively high. It would be great if the state value can be estimated relatively accurately through some mechanism and used to guide the strategy update. The Actor-Critic algorithm does just that, and the algorithm is divided into two parts: Actor and Critic. Among them, $Actor$ is responsible for updating the strategy, and $Critic $ is responsible for updating the value. When Critic updates the value, it is actually the $Q-learning $ algorithm or the $SARSA $\footnote{State–action–reward–state–action (SARSA) } algorithm.

The pseudo code of Actor-Critic policy is as follow:

\begin{figure}[H] 
\centering 
\includegraphics[width=0.7\textwidth]{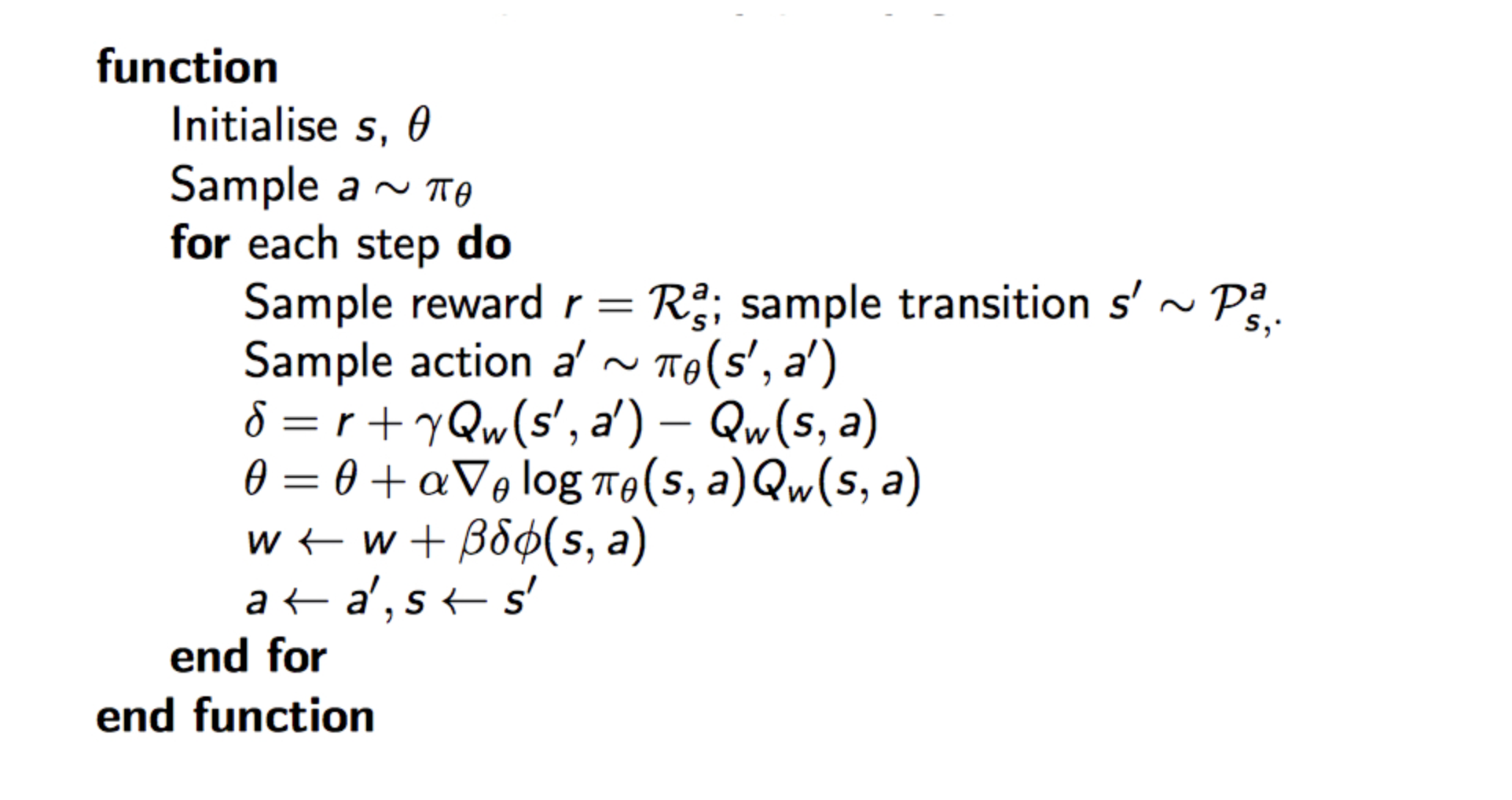} 
\label{Fig.main2} 
\end{figure}

Let me explain the $actor-critic$ method in general.
I use neural networks as an example; in fact, you can use linear functions, $kernel$ and other methods to approximate functions.
$Actor$(: In order to play this game and get the highest reward possible, you need to implement a function: input state, output action, which is the second step above. You can use neural networks to approximate this function. The remaining task is how to Train the neural network to make it perform better (higher reward). This network is called an actor
Critic (Jury): In order to train an actor, you need to know how the actor is performing, and decide on the adjustment of the neural network parameters based on the performance. This requires the use of $"Q-value"$ in reinforcement learning. But $Q-value$ is also an unknown function, so it can also be approximated by a neural network. This network is called critic.

Actor-Critic training.
Let me explain in general terms.
The Actor sees the current state of the game and makes an action.
Critic scores the actor's performance just now based on both state and action.
Actor adjusts its strategy (actor neural network parameters) according to the score of the critic (jury), and strives to do better next time.
Critic adjusts its scoring strategy (critic neural network parameters) based on the reward (equivalent to ground truth) given by the system and the scores of other judges (critic targets).
In the beginning, $actor$ performed randomly, and critic scored randomly. However, due to the existence of $reward$, $critic$ scores are getting more accurate, and $actor$ performs better and better.

It feels like GAN\footnote{Generative  adversarial network} , the two networks are colliding with each other.

\section{Code Implementation}

\paragraph{The code for Monte Carlo Policy Gradient.}

The black trolley is connected to a movable rod through a bearing and moves on a frictionless track. The movement is completed by applying a thrust of +1 or -1 to the trolley. If the trolley stops, the wooden pole will definitely fall down. The experiment ensures that the wooden pole stands up by controlling the trolley to move left and right. If the left and right angles of the wooden pole are greater than 15 degrees or the cart moves more than 2.4 units left and right, the game ends immediately. The wood stick is saved and erected in each time step, and it is rewarded with +1.

Next, we load the CartPole-v1 environment through Gym. Since Notebook does not support the visualization of the CartPole-v1 environment, rendering operations cannot be performed.

\begin{python}
#
import gym
import warnings
warnings.filterwarnings('ignore')

env = gym.make('CartPole-v1')
\end{python}

We try to use random actions to see how many time steps each episode can last:

\begin{python}
#
for _ in range(10):
    t = 0
    env.reset()
    while True:
        action = env.action_space.sample()
        observation, reward, done, _ = env.step(action)
        t += 1
        if done:
            print("Episode finished after {} timesteps".format(t))
            break
\end{python}

You can find that under random actions, each Episode generally does not exceed 50 time steps. The time step here actually corresponds to the cumulative reward value of the episode.

Next, we implement the Monte Carlo strategy gradient-based reinforcement learning algorithm according to the Monte Carlo strategy gradient algorithm pseudo code provided above.

\begin{python}
#
def mc_policy_gradient(env, theta, lr, episodes):
    """
    Parameter:
    env -- environment
    theta -- parameter
    lr -- learning rate
    episodes -- iteration times

    Return: 
    episodes -- parameter
    """
    # Monte-Carlo Policy Gradient
    for episode in range(episodes):  # iterate episode
        episode = []
        start_observation = env.reset()  # initised the environment
        t = 0
        while True:
            policy = np.dot(theta, start_observation)  # compute the policy value
            # in this case, action_space is 2, using Sigmoid function to deal with policy value
            pi = 1 / (1 + np.exp(-policy))
            if pi >= 0.5:
                action = 1  # push right
            else:
                action = 0  # push left
            next_observation, reward, done, _ = env.step(action)  # take action
            # the value returned from environment add into episode
            episode.append([next_observation, action, pi, reward])
            start_observation = next_observation 
            # put the return value of observation as the next obervation of iteration
         
            t += 1
            if done:
                print("Episode finished after {} timesteps".format(t))
                break
        #update the parameter in terms of last episode theta
      
        for timestep in episode:
            observation, action, pi, reward = timestep
            theta += lr * (1 - pi) * np.transpose(-observation) * reward

    return theta

\end{python}

\begin{python}

for _ in range(10):
    t = 0
    env.reset()
    while True:
        action = env.action_space.sample()
        observation, reward, done, _ = env.step(action)
        t += 1
        if done:
            print("Episode finished after {} timesteps".format(t))
            break

\end{python}

\begin{python}
#
import numpy as np
lr = 0.02
theta = np.random.rand(4)
episodes = 10

mc_policy_gradient(env, theta, lr, episodes
\end{python}

It can be found that with the introduction of the Monte Carlo strategy gradient algorithm, each Episode will generally reach hundreds of time steps (by default, $max_episode_steps$ = 500). Of course, due to the random initialization of the parameters, sometimes a good learning result cannot be obtained. However, if you re-pass the parameters of a relatively good result into the function, you can get a better result almost every time, which is the process of continuous reinforcement learning.

\paragraph{The code for Actor-Critic Policy Gradient}

In order to make life easier,We choose Keras framework to implement neural networks.

\begin{python}
#
import gym
import numpy as np
import tensorflow as tf
from tensorflow import keras
from tensorflow.keras import layers

# Configuration paramaters for the whole setup
seed = 42
gamma = 0.99  # Discount factor for past rewards
max_steps_per_episode = 10000
env = gym.make("CartPole-v0")  # Create the environment
env.seed(seed)
eps = np.finfo(np.float32).eps.item()  # Smallest number such that 1.0 + eps != 1.0
\end{python}


\begin{python}
#
num_inputs = 4
num_actions = 2
num_hidden = 128

inputs = layers.Input(shape=(num_inputs,))
common = layers.Dense(num_hidden, activation="relu")(inputs)
action = layers.Dense(num_actions, activation="softmax")(common)
critic = layers.Dense(1)(common)

model = keras.Model(inputs=inputs, outputs=[action, critic])

\end{python}

\begin{python}
optimizer = keras.optimizers.Adam(learning_rate=0.01)
huber_loss = keras.losses.Huber()
action_probs_history = []
critic_value_history = []
rewards_history = []
running_reward = 0
episode_count = 0

while True:  # Run until solved
    state = env.reset()
    episode_reward = 0
    with tf.GradientTape() as tape:
        for timestep in range(1, max_steps_per_episode):
            # env.render(); Adding this line would show the attempts
            # of the agent in a pop up window.

            state = tf.convert_to_tensor(state)
            state = tf.expand_dims(state, 0)

            # Predict action probabilities and estimated future rewards
            # from environment state
            action_probs, critic_value = model(state)
            critic_value_history.append(critic_value[0, 0])

            # Sample action from action probability distribution
            action = np.random.choice(num_actions, p=np.squeeze(action_probs))
            action_probs_history.append(tf.math.log(action_probs[0, action]))

            # Apply the sampled action in our environment
            state, reward, done, _ = env.step(action)
            rewards_history.append(reward)
            episode_reward += reward

            if done:
                break

        # Update running reward to check condition for solving
        running_reward = 0.05 * episode_reward + (1 - 0.05) * running_reward

        # Calculate expected value from rewards
        # - At each timestep what was the total reward received after that timestep
        # - Rewards in the past are discounted by multiplying them with gamma
        # - These are the labels for our critic
        returns = []
        discounted_sum = 0
        for r in rewards_history[::-1]:
            discounted_sum = r + gamma * discounted_sum
            returns.insert(0, discounted_sum)

        # Normalize
        returns = np.array(returns)
        returns = (returns - np.mean(returns)) / (np.std(returns) + eps)
        returns = returns.tolist()

        # Calculating loss values to update our network
        history = zip(action_probs_history, critic_value_history, returns)
        actor_losses = []
        critic_losses = []
        for log_prob, value, ret in history:
            # At this point in history, the critic estimated that we would get a
            # total reward = `value` in the future. We took an action with log probability
            # of `log_prob` and ended up recieving a total reward = `ret`.
            # The actor must be updated so that it predicts an action that leads to
            # high rewards (compared to critic's estimate) with high probability.
            diff = ret - value
            actor_losses.append(-log_prob * diff)  # actor loss

            # The critic must be updated so that it predicts a better estimate of
            # the future rewards.
            critic_losses.append(
                huber_loss(tf.expand_dims(value, 0), tf.expand_dims(ret, 0))
            )

        # Backpropagation
        loss_value = sum(actor_losses) + sum(critic_losses)
        grads = tape.gradient(loss_value, model.trainable_variables)
        optimizer.apply_gradients(zip(grads, model.trainable_variables))

        # Clear the loss and reward history
        action_probs_history.clear()
        critic_value_history.clear()
        rewards_history.clear()

    # Log details
    episode_count += 1
    if episode_count 
        template = "running reward: {:.2f} at episode {}"
        print(template.format(running_reward, episode_count))

    if running_reward > 195:  # Condition to consider the task solved
        print("Solved at episode {}!".format(episode_count))
        break

\end{python}

\section{Comparison between two policies}

\begin{figure}[htbp]
\centering
\begin{minipage}[t]{0.48\textwidth}
\centering
\includegraphics[width=6cm]{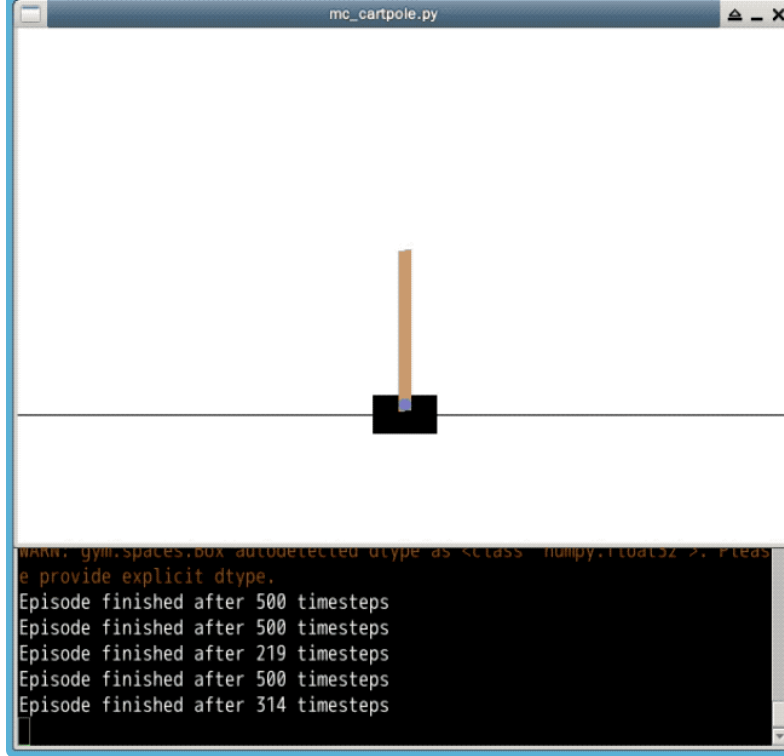}
\caption{The result for Monte Carlo  Policy}
\end{minipage}
\begin{minipage}[t]{0.48\textwidth}
\centering
\includegraphics[width=6cm]{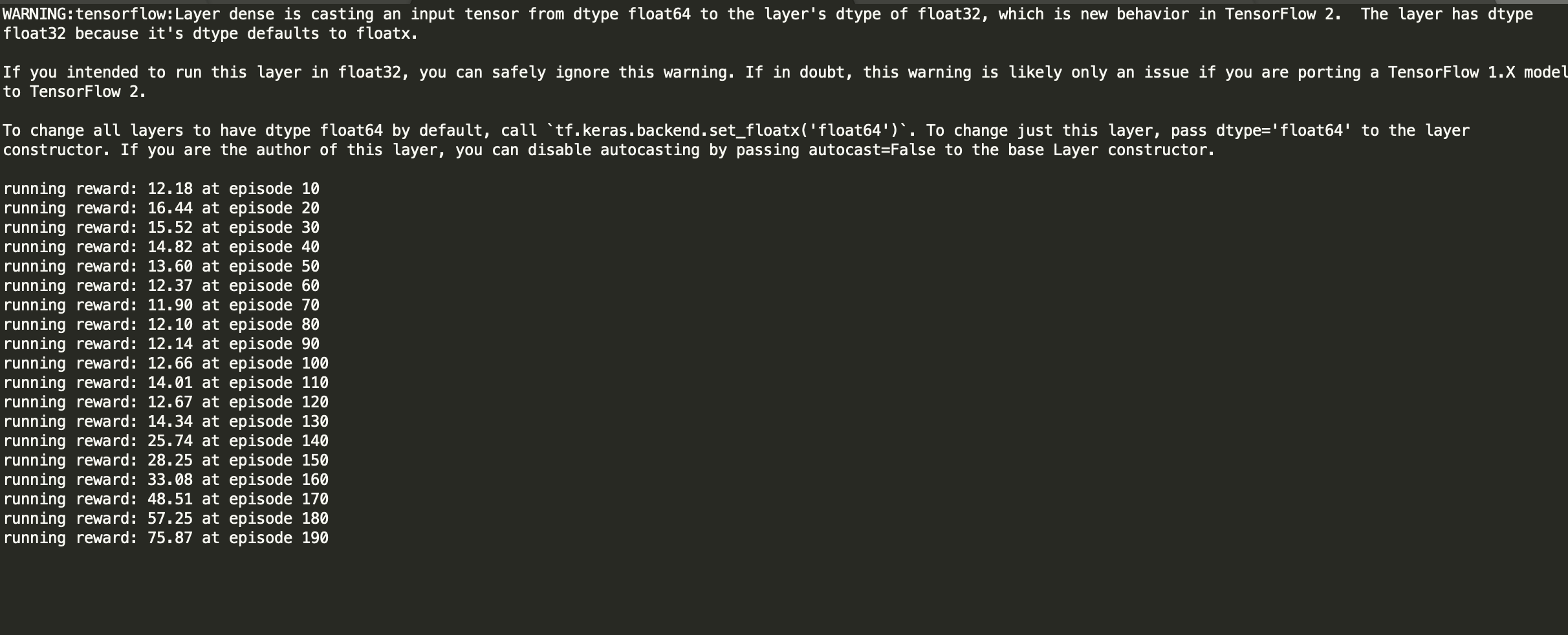}
\caption{The result for Actor-Critic Policy}
\end{minipage}
\end{figure}
\paragraph{Observation}
As can be seen from the below two figures, the Monte Carlo strategy requires many iterations and the variance is relatively large,since the Monte carol policy would experience 500 episode before coverage,whereas the actor-critic only needs 190 episode.
\paragraph{Explanation}
The Monte Carlo strategy gradient method uses the harvest as the estimation of the state value. Although it is unbiased, the noise is relatively large, that is, the variability (variance) is relatively high. If we can estimate the state value relatively accurately and use it to guide policy updates, will there be better learning effects? This is the main idea of the Actor-Critic strategy gradient. Among them, Critic is used to estimate the value of behavior:

$$   
Q_{w}(s, a) \approx Q^{\pi_{\theta}}(s, a)
$$

Gradient policy based on Actor-Critic  is divided into two parts:

1 Critic: Parameterize behavioral $value function$ $Q_{w}(s, a)$, and then update the weight.

2 Actor: Navigate the update of the policy function parameter $\theta$ according to the value obtained in the $Critic $part.

\section{ Expansibility to F1tenth Autonomous cars}

\begin{figure}[htbp]
\centering
\begin{minipage}[t]{0.48\textwidth}
\centering
\includegraphics[width=6cm]{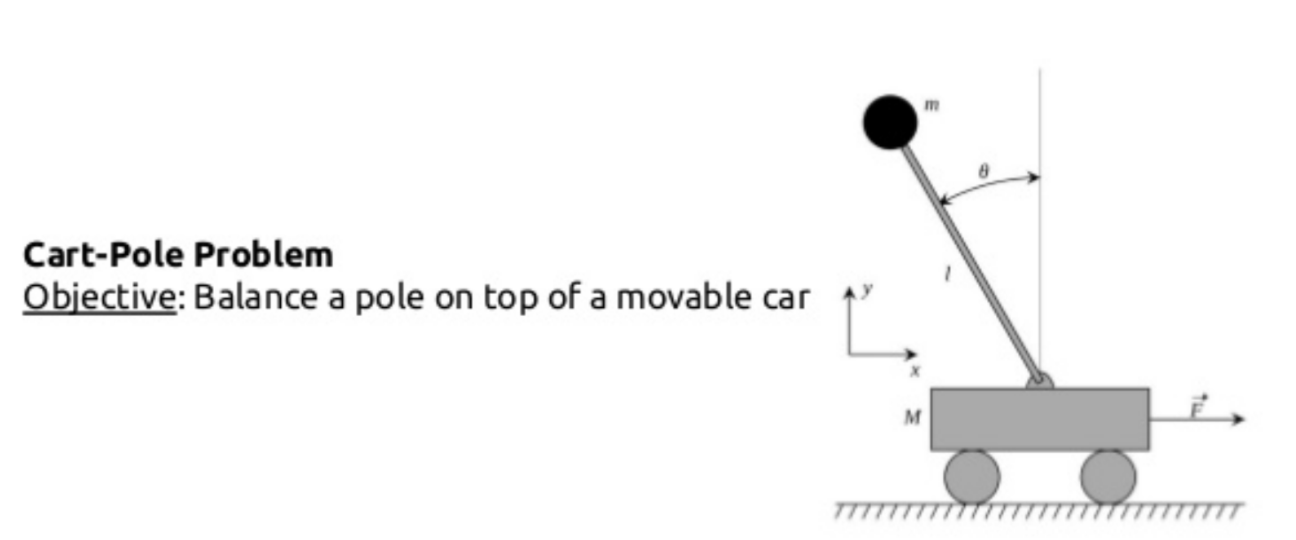}
\caption{The model for CartPole}
\end{minipage}
\begin{minipage}[t]{0.48\textwidth}
\centering
\includegraphics[width=6cm]{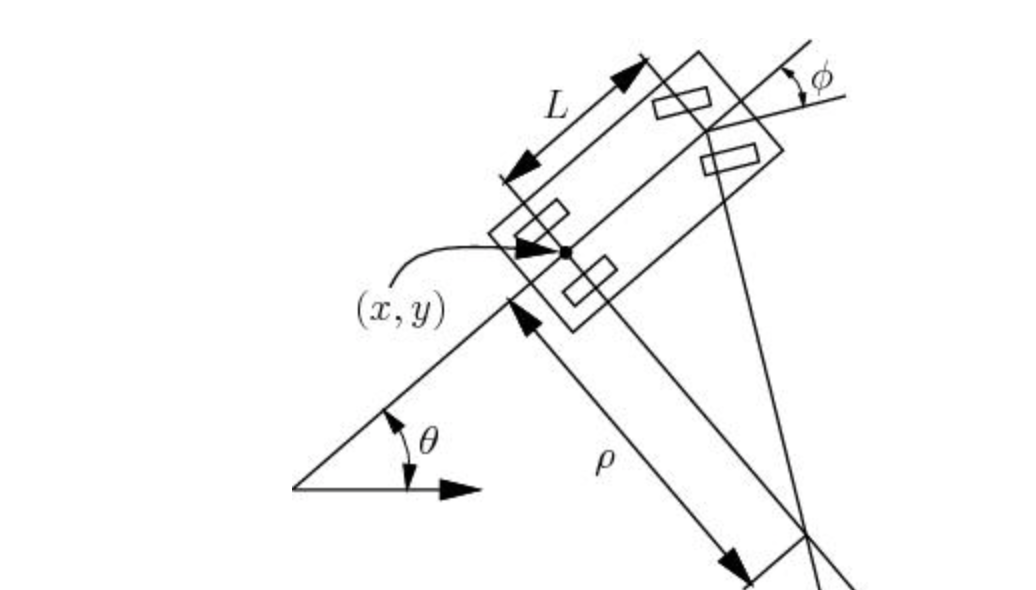}
\caption{The model for F1tenth vehicles}
\end{minipage}
\end{figure}

\paragraph{Vehicle kinematics model in autonomous driving}

To control the movement of the vehicle, we must first establish a numerical model of the movement of the vehicle. The more accurate the model is, the more accurate the description of the movement of the vehicle is, and the better the effect of tracking and controlling the vehicle. In addition to truly reflecting the characteristics of the vehicle, the established model should also be as simple and easy to be used as possible. $Bicycle Model $is a generic vehicle kinematics model.

The establishment of the $Bicycle Model $is based on the following assumptions:

1) The movement of the vehicle in the vertical direction (Z-axis direction) is not considered, that is, it is assumed that the movement of the vehicle is a movement on a two-dimensional plane.

2) Assume that the left and right tires of the vehicle have the same steering angle and speed at any time; in this way, the movement of the left and right tires of the vehicle can be combined into one time to describe.

3) It is considered that the speed of the vehicle changes slowly, and the transfer of front and rear axle loads is ignored.

4) It is considered that both the body and suspension system are rigid systems.

5) It is assumed that the movement and steering of the vehicle are driven by the front wheels (front-wheel-only).

In order to clearly compare the similarities between CartPole's model and the autonomous vehicle model, we use a simple model to illustrate this problem.

\paragraph{Vehicle motion model with the rear axis as the origin}

We could see this model on Figure 4.

The Autonomous vehicle model  can be simplified to a rigid body structure moving on a two-dimensional plane. The state of the vehicle at any time $q=(x, y, \theta)$, the origin of the vehicle coordinates is at the center of the rear axle, and the coordinate axis is parallel to the vehicle body. $s$ represents the speed of the vehicle, $\phi$ formula] represents the steering angle (left is positive, right is negative), L represents the distance between the front wheel and the rear wheel, if the steering angle remains  $\phi$  unchanged, the vehicle will make a circle on the spot with the radius is $\rho$

Vehicle motion model:

$$   
\dot{x}=f_{1}(x, y, \theta, s, \phi)
$$

$$
\dot{y}=f_{2}(x, y, \theta, s, \phi)
$$

$$
\dot{\theta}=f_{3}(x, y, \theta, s, \phi)
$$

In a very short time $\Delta t$, it can be approximately considered that the vehicle is moving in the direction of the body. $d_{x}, \quad d_{y}$ indicate the distance the vehicle moves on the x-axis and y-axis in the $d_{t}$ time 

As the following formulas:
$$
\frac{d y}{d x}=\tan \theta
$$
$$
\frac{d y}{d x}=\frac{\dot{y}}{\dot{x}}
$$
$$
\tan \theta=\sin \theta / \cos \theta
$$

We could derive:
$$
-\dot{x} \sin \theta+\dot{y} \cos \theta=0
$$

Use $\omega$ to express the distance that the body moves in the time $\Delta t$, so by:

$$
d \omega=\rho d \theta
$$

$$\rho=L / \tan \phi$$

It could be derived:
$$d \theta=\frac{\tan \phi}{L} d \omega
$$

Divide both sides of the equation by [$d_{t}$ and based on condition  on $\dot{\omega}=s$,we can finally get the equation:

$$\dot{\theta}=\frac{s}{L} \tan \phi$$

At this point, a simplified non-holonomically constrained vehicle motion model is completed, which is summarized as follows:

$$
\left[\begin{array}{c}\dot{x} \\ \dot{y} \\ \dot{\theta}\end{array}\right]=s *\left[\begin{array}{c}\cos (\theta) \\ \sin (\theta) \\ \frac{\tan \phi}{L}\end{array}\right]
$$

$$
\dot{s}=a
$$

Based on this simple kinematics model, given the control input $(a, \phi)$ at a certain moment, we can estimate the state information (coordinates, yaw angle and speed) of the vehicle at the next moment.

\paragraph{Conclusion for Similarity}

As can be seen from the above figures and explanation, the rotation angle of CartPole is similar to the angle of the F1tenth Autonomous vehicle when turning. Because these are the parameters that our strategy needs to learn during reinforcement learning.
They might find a balance and eventually converge. We can make a conjecture that the model in this paper can be taken in order to Autonomous vehicles when they are turning in trajectory.

\end{document}